\documentclass[conference]{IEEEtran}
\usepackage{times}

\usepackage[numbers]{natbib}
\usepackage{multicol}
\usepackage[bookmarks=true]{hyperref}
\usepackage{graphicx}
\usepackage{multirow}
\usepackage{mymacros}

\newtheorem{definition}{Definition}

\begin{document}

\title{Chance-Constrained Trajectory Optimization for High-DOF Robots in Uncertain Environments}


\author{\authorblockN{Charles Dawson, Ashkan Jasour, Andreas Hofmann, Brian Williams}
\authorblockA{Department of Aeronautics and Astronautics\\
Massachusetts Institute of Technology\\
Cambridge, Massachusetts 02139--4301\\
Email: \{cbd, jasour, hofma, williams\}@mit.edu}}


%

\maketitle

\begin{abstract}
Many practical applications of robotics require systems that can operate safely despite uncertainty. In the context of motion planning, two types of uncertainty are particularly important when planning safe robot trajectories. The first is environmental uncertainty --- uncertainty in the locations of nearby obstacles, stemming from sensor noise or (in the case of obstacles' future locations) prediction error. The second class of uncertainty is uncertainty in the robots own state, typically caused by tracking or estimation error. To achieve high levels of safety, it is necessary for robots to consider both of these sources of uncertainty. In this paper, we propose a risk-bounded trajectory optimization algorithm, known as Sequential Convex Optimization with Risk Optimization (SCORA), to solve chance-constrained motion planning problems despite both environmental uncertainty and tracking error. Through experiments in simulation, we demonstrate that SCORA significantly outperforms state-of-the-art risk-aware motion planners both in planning time and in the safety of the resulting trajectories.
\end{abstract}

\IEEEpeerreviewmaketitle

\section{Introduction}

For most robots, developed in the cradle of a well-controlled, carefully structured simulation or laboratory environment, primary challenge posed by deployment in the outside world is \textit{uncertainty}. Imagine a robot making a delivery on a factory floor, or an autonomous vehicle driving on a busy road. A nearby human might turn left at an upcoming intersection, but they could just as easily turn right, cutting across the robot's planned path. The robot might have a camera to detect obstacles in its way, but its obstacle-detection algorithms may be error-prone. When it comes time to execute a planned path, the robot may only be able to track its intended path to within some tolerance, leading to uncertainty in its own state.

All of these factors point to a need for robots than can operate safely despite uncertainty both in the state of the environment (e.g. robustness to perception and prediction errors) and in the state of the robot (e.g. robustness to estimation and tracking errors). There is a large body of work dealing with chance-constrained motion planning --- which aims to find trajectories where the probability of failure (i.e. collision) is below some bound --- when there is uncertainty only in the state of the robot \cite{masahiroonoIterativeRiskAllocation2008,blackmoreConvexChanceConstrained2009,vandenbergLQGMPOptimizedPath2011,daiChanceConstrainedMotion2018,ludersChanceConstrainedRRT2010}. Similarly, there are a number of techniques for finding safe paths when the state of the external environment is uncertain \cite{axelrodProvablySafeRobot2018,jasourRiskContoursMap2019,parkFastBoundedProbabilistic2018,dawsonProvablySafeTrajectory2020}. However, relatively little work has been done to build a chance-constrained motion planner that simultaneously considers the risk due to environmental uncertainty and tracking error uncertainty. Additionally, many of the existing approaches on chance-constrained motion planning are restricted to highly simplified geometric models. Some planners restrict themselves to point-mass models, which cannot represent high degree-of-freedom robots such as manipulators \cite{masahiroonoIterativeRiskAllocation2008,blackmoreConvexChanceConstrained2009,ludersChanceConstrainedRRT2010,axelrodProvablySafeRobot2018,jasourRiskContoursMap2019}. Others rely on collections of many small spheres to ``bubble-wrap'' the scene geometry, needlessly increasing computational cost when modeling objects such as humans or furniture \cite{parkFastBoundedProbabilistic2018}. Other approaches can model more complicated geometry (such as \cite{daiChanceConstrainedMotion2018}) but are slow to find solutions and prone to underestimating the risk of collision.

As a result of these gaps in the state of the art, there is an unmet need for a chance-constrained motion planner that can simultaneously consider uncertainty in both the environment and robot state even when the robot and environment are modeled using general convex geometry.

\subsection{Contributions}

In this paper, we present a chance-constrained trajectory optimization algorithm that is capable of meeting all of these needs. This algorithm, called Sequential Convex Optimization with Risk Allocation (SCORA) successfully manages risk due to both environmental uncertainty and tracking error; supports robots and environments with rich, convex geometry; and quickly finds high-quality trajectories that limit the risk of collision between the robot and its environment.

The core of this algorithm are differentiable risk estimates known as $\epsilon$-shadows, which we extend from previous work to provide robustness to state uncertainty and tracking error as well as safety in the face of environmental uncertainty. Using these estimates, SCORA employs a novel approach to chance-constrained non-convex optimization by solving a sequence of chance-constrained convex approximations to the full non-convex trajectory planning problem. As we demonstrate in this paper, our approach significantly outperforms comparable algorithms in both run-time and solution quality.

\section{Related work}

Early works on chance-constrained motion planning focused on the simple case of a point robot navigating a convex space, where the user has specified a maximum acceptable probability of violating any of the convex constraints on the robot \cite{vanhessemStochasticInequalityConstrained2004,masahiroonoIterativeRiskAllocation2008,blackmoreConvexChanceConstrained2009}. Although the applicability of these planners is limited to relatively simple environments, their key insight, that more efficient paths may be found by dynamically allocating risk between constraints, is broadly applicable to chance-constrained planning problems. This insight gave rise to the iterative risk allocation (IRA \cite{masahiroonoIterativeRiskAllocation2008}) and convex risk allocation  (CRA \cite{blackmoreConvexChanceConstrained2009}) algorithms. IRA works by repeatedly solving an inner-loop optimization problem and changing the allocation of risk between constraints at each step, while CRA folds risk allocation into the inner-loop optimization (increasing the problem complexity from linear programming to convex programming but avoiding a costly outer-loop optimization). CRA was later extended to employ mixed-integer convex optimization to solve problems involving a point robot navigating around polytope obstacles \cite{blackmoreChanceConstrainedOptimalPath2011}. A more recent work known as p-Chekov \cite{daiChanceConstrainedMotion2018} employs IRA as an outer-loop around non-convex trajectory optimization to handle robots with non-trivial geometry, but this approach is limited by long run-times and inaccurate risk estimation. Moreover, none of these techniques are designed to consider uncertainty in the configuration of obstacles, although p-Chekov can be extended to handle environmental uncertainty as well as tracking error (we will use this extension as a main point of comparison for the performance of our approach). As we discuss in the following sections, our approach builds on the foundation of CRA, which we extend to support both environmental uncertainty and tracking error, including risk allocation within a non-convex optimization problem.

In parallel to these optimization-based works, a number of techniques extend traditional sampling-based motion planning algorithms such as PRM and RRT to the chance-constrained context \cite{bryRapidlyexploringRandomBelief2011,axelrodProvablySafeRobot2018,ludersChanceConstrainedRRT2010}. Most of these techniques are limited to point-mass representations of robot geometry, but they add important developments to the theory of chance-constrained motion planning. In particular, Axelrod, Kaelbling, and Lozano-P\'erez in \cite{axelrodProvablySafeRobot2018} introduce the notion of $\epsilon$-shadows: geometric objects that provide a means of quickly computing upper-bounds on the risk of collision due to environmental uncertainty. Dawson \textit{et al.} extend the theory of $\epsilon$-shadows in \cite{dawsonProvablySafeTrajectory2020} to support arbitrary convex robot geometry and derived the gradient of $\epsilon$-shadow risk estimates with respect to robot state. Dawson \textit{et al.} use this framework to develop an efficient trajectory optimization algorithm that handles both rich geometry and environmental uncertainty; our work in this paper can be seen as extending this approach to handle state uncertainty as well. We will present the necessary theoretical background for $\epsilon$-shadow-based trajectory optimization in Section~\ref{iros-recap}.

\section{Problem statement}

We consider the problem of trajectory optimization over a fixed, finite horizon $T$: finding a sequence of nominal states $\bar{q}_0, \bar{q}_1, \ldots, \bar{q}_T \in \R^n$ that navigate between starting and final configurations $\mathcal{Q}_{start}$ and $\mathcal{Q}_{final}$ while limiting the risk of collision incurred during the motion. To model state uncertainty, we assume that the our decision variables specify the nominal path $\bar{q} = \mat{\bar{q}_0, \bar{q}_1, \ldots, \bar{q}_T}$ and that at execution-time the robot will follow some realization of this trajectory $q$ drawn from a multivariate Gaussian distribution about the nominal $\bar{q}$. That is, $q \sim \mathcal{N}(\bar{q}, \Sigma_q)$. For notational convenience, we consider the distribution of the full trajectory (the concatenation of the state at each waypoint), which allows us to easily consider cases when the tracking error $q - \bar{q}$ is correlated across time, as might be the case when employing linear quadratic Gaussian (LQG) control \cite{vandenbergLQGMPOptimizedPath2011}. We assume that an estimate of the tracking error covariance $\Sigma_q$ is known a-priori, as is the case when the system is controlled using LQG.

We model the environment $E$ as a set of convex obstacles $\mathcal{O}$; to represent environmental uncertainty we assume that each ground truth obstacle $\mathcal{O}_i$ is offset from a known, nominal obstacle $O_i$ by some uncertain translation $d\sim\mathcal{N}(0, \Sigma_{O})$. To model collisions, we consider the signed distance function $\textrm{sd}_{\mathcal{O}}(q)$, which returns the minimum distance from the robot to an obstacle $\mathcal{O} \in E$. By definition, $\textrm{sd}_{\mathcal{O}}(q) > 0$ if the robot is not in collision with $\mathcal{O}$, and $\textrm{sd}_{\mathcal{O}}(q) < 0$ if the robot is in collision (in this case, the signed distance is equal to the distance by which the robot penetrates the obstacle, expressed as a negative quantity).

Given a user-specified limit on the overall risk of collision $\Delta$, an arbitrary convex cost function $f$ (e.g. a quadratic cost on displacement between timesteps), and a set of arbitrary inequality constraints $g_i$ (e.g. enforcing joint limits or plant dynamics), the chance-constrained trajectory optimization problem is:

\begin{align}
    \min_{\bar{q}_0, \bar{q}_1, \ldots, \bar{q}_T} \quad & f(\bar{q}_0, \bar{q}_1, \ldots, \bar{q}_T) \tag{ccNLP-1}\label{full_ccNLP}\\
    \text{s.t.} \qquad & \bar{q}_0 \in \mathcal{Q}_{start};\ \bar{q}_T \in \mathcal{Q}_{final} \\
    & \textrm{Pr}_{\substack{q \sim \cN(\bar{q}, \Sigma_q) \\ \mathcal{O} = O + d;\ d\sim\cN(0, \Sigma_O)}}\pn{\bigwedge_{0 \leq t \leq T} \bigwedge_{\mathcal{O} \in E} \textrm{sd}_{\mathcal{O}}(q_t) \geq 0} \nonumber\\
    &\qquad\qquad\qquad\qquad\qquad\qquad\qquad \geq 1-\Delta \label{full_ccNLP_chance_constraint}\\
    & g_i(\bar{q}_0, \bar{q}_1, \ldots, \bar{q}_T) \leq 0;\quad i \in \mathcal{I}
\end{align}

When the robot and environment are both modeled as collections of convex shapes, the signed distance can be computed easily using standard computational geometry algorithms \cite{coumansBulletPhysicsEngine}. Unfortunately, the probability of the signed distance dropping below zero, i.e. the probability of collision which we wish to constrain in \eqref{full_ccNLP_chance_constraint}, is not available in closed form. Instead, this probability must be estimated in order to yield a tractable optimization problem. In Section~\ref{iros-recap}, we will discuss how this probability can be estimated in the presence of environmental uncertainty alone (drawing on previous work in this area), and in Section~\ref{state-uncertainty} we make our main contribution by expanding our view to develop an optimization-based algorithm that efficiently manages risk due to both environmental uncertainty and tracking error. Finally, in Section~\ref{results}, we will present empirical results that demonstrate how considering both types of uncertainty provides a significantly higher level of safety than considering environmental uncertainty alone.

\section{Safety in uncertain environments}\label{iros-recap}

The probability of collision constraining problem~\eqref{full_ccNLP} in \eqref{full_ccNLP_chance_constraint} captures the risk stemming from uncertainty both in state tracking error $q \sim \cN(\bar{q}, \Sigma_q)$ and in the locations of obstacles in the environment $\mathcal{O} = O + d;\ d\sim\cN(0, \Sigma_O)$. A sound approach to chance-constrained motion planning must consider both sources of uncertainty, but it is conceptually more straightforward to first isolate the effects of environmental uncertainty (this section) then expand our approach to consider state uncertainty (Section~\ref{state-uncertainty}).

Considering only environmental uncertainty (i.e. assuming that $q = \bar{q}$), the probability in constraint~\eqref{full_ccNLP_chance_constraint} can be rewritten using Boole's inequality:
\begin{align}
    \textrm{Pr}_{d\sim\cN(0, \Sigma_O)}\pn{\bigwedge_{0 \leq t \leq T} \bigwedge_{\mathcal{O} \in E} \textrm{sd}_{\mathcal{O}}(q_t) \geq 0}&\geq 1-\Delta \\
    \textrm{Pr}_{d\sim\cN(0, \Sigma_O)}\pn{\bigvee_{0 \leq t \leq T} \bigvee_{\mathcal{O} \in E} \textrm{sd}_{\mathcal{O}}(q_t) \leq 0} &\leq \Delta\\
    \sum_{0 \leq t \leq T} \sum_{\mathcal{O} \in E} \textrm{Pr}_{d\sim\cN(0, \Sigma_O)}\pn{\textrm{sd}_{\mathcal{O}}(q_t) \leq 0} &\leq \Delta \label{reduced_chance_constraint}
\end{align}

This reduces the problem of evaluating the risk of collision across the entire trajectory to the simpler problem of bounding the risk of collision between in a specific configuration $q_t$. To bound this probability, we follow \cite{axelrodProvablySafeRobot2018} and more recently \cite{dawsonProvablySafeTrajectory2020} in using $\epsilon$-shadows to estimate an upper bound on this risk.

\begin{definition}[$\epsilon$-shadow]
A set $\mathcal{S} \subset \R^3$ is a maximal $\epsilon$-shadow of an uncertain obstacle $\mathcal{O}$ if the probability $\textrm{Pr}\pn{\mathcal{O} \subseteq S} = 1-\epsilon$.
\end{definition}

Intuitively, an $\epsilon$-shadow is a geometric object (often an enlarged version of the nominal obstacle) that contains its associated uncertain obstacle with probability $1-\epsilon$. As a result, the $\epsilon$-shadow acts as a mathematically rigorous safety buffer: if the robot avoids collision with the $\epsilon$-shadow then it necessarily limits the risk of collision with the obstacle itself to less than $\epsilon$. As a result, we can simplify the problem of computing an upper bound on the risk of collision between the robot an an obstacle to simply finding the smallest $\epsilon$ such that there is no collision between the robot and the corresponding $\epsilon$-shadow.

\begin{figure}[tbp]
    \centering
    \includegraphics[width=0.9\linewidth]{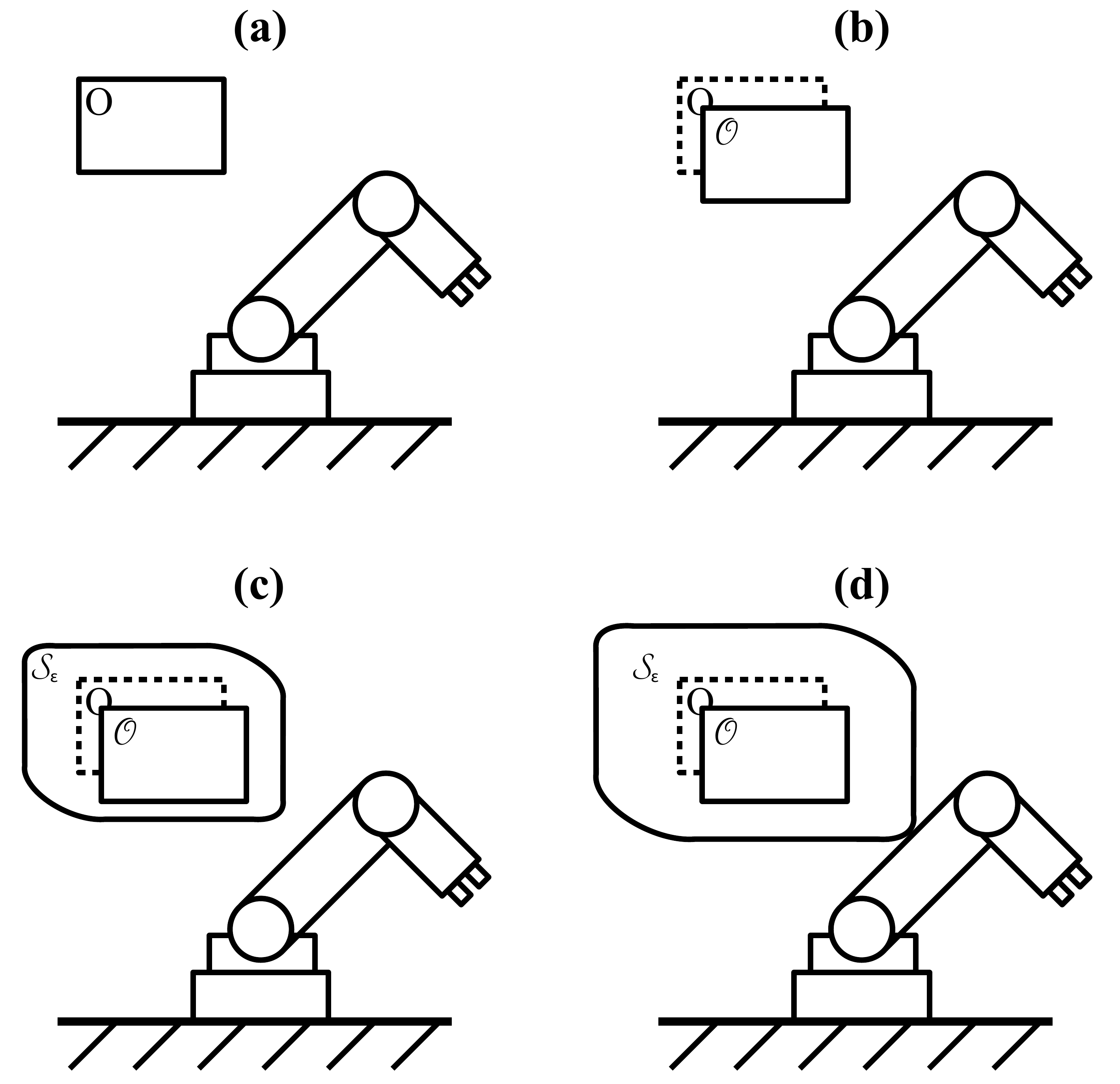}
    \caption{(a) $\epsilon$-shadows are used to compute an upper bound on the risk of collision between an obstacle $O$ and the robot in some configuration. (b) We assume that the nominal geometry and location of the obstacle is known, but that the true location of the obstacle is uncertain. (c) Given some $\epsilon$, we construct an $\epsilon$-shadow by asymmetrically expanding the nominal geometry $O$ in the directions in which the obstacle position is most uncertain; the true obstacle $\mathcal{O}$ is guaranteed to lie within the shadow $\mathcal{S}_{\epsilon}$ with probability $1-\epsilon$. (d) The $\epsilon$-shadow is convex and can be checked for collision with the robot in linear time, so we apply an bisection line search in $\epsilon$ to efficiently find the smallest $\epsilon$ that upper bounds the risk of collision with the obstacle (corresponding to the largest $\epsilon$-shadow that does not intersect the robot).}
    \label{fig:shadow_explainer}
\end{figure}

The process of finding these $\epsilon$-shadows is summarized in Figure~\ref{fig:shadow_explainer}. Essentially, to construct the shadow at a given risk level $\epsilon$, we take the Minkowski sum of the nominal obstacle geometry with an ellipsoid: $\mathcal{S}_{\epsilon} = O \bigoplus \mathcal{D}_{\epsilon}$, where $\mathcal{D}_{\epsilon} = \set{x\ :\ x^T \Sigma_O^{-1} x \leq \phi^{-1}(1-\epsilon)}$, $\Sigma_{O}$ is the covariance matrix for the uncertainty in the obstacle's location, and $\phi^{-1}$ is the inverse cumulative distribution function of the $\chi^2$ distribution with 3 degrees of freedom. Intuitively, the boundary of $\mathcal{D}_{\epsilon}$ is an isoprobability surface (at probability $1-\epsilon$) of the uncertain translation of the true obstacle $\mathcal{O}$, and so the sum $\mathcal{S}_{\epsilon} = O \bigoplus \mathcal{D}_{\epsilon}$ contains all possible translations of $O$ with probability $1-\epsilon$ as well. Our focus in this paper is not to reproduce the proofs of the correctness of $\epsilon$-shadow risk estimates (those proofs can be found in \cite{dawsonProvablySafeTrajectory2020}); instead, we will focus in Section~\ref{state-uncertainty} on how these $\epsilon$-shadows can be combined with a robust non-convex optimization strategy to plan safe trajectories in the presence of both environmental uncertainty and tracking error.

These $\epsilon$-shadows have two important properties. First, if the underlying geometry $O$ is convex, then the shadow $\mathcal{S}_{\epsilon}$ is also convex (since ellipsoids are convex the Minkowski sum of two convex shapes is also convex). Second, modern computational geometry libraries allow the Minkowski sum of two convex shapes to be represented implicitly via their support vectors \cite{coumansBulletPhysicsEngine}. These two properties mean that we can construct an $\epsilon$-shadow and check for collision between the shadow and the robot very quickly (the time complexity is linear in the number of vertices in the obstacle and robot geometry \cite{gilbertFastProcedureComputing1988}). Furthermore, beyond simply checking whether the robot is safe at any fixed risk level, Dawson \textit{et al}. demonstrate that a simple line search algorithm can be used to compute the \textit{largest} convex $\epsilon$-shadow that does not collide with the robot in state $q_t$ \cite{dawsonProvablySafeTrajectory2020}. This largest $\epsilon$-shadow corresponds to the smallest $\epsilon$, i.e. the least upper bound on the risk of collision between the robot and the obstacle at that state. Computing this tight upper bound on collision probability, which we denote as $\epsilon_\mathcal{O}(q_t)$ (with respect to one obstacle $O$ and a particular robot configuration $q_t$), can be done on the order of \SI{100}{\micro s}. Using this estimate, we can provide a deterministic inner approximation of the probabilistic constraint \eqref{reduced_chance_constraint}:

\begin{align}
    \sum_{0 \leq t \leq T} \sum_{\mathcal{O} \in E} \epsilon_{\mathcal{O}}(q_t) &\leq \Delta \label{deterministic_e_shadow_constrain}
\end{align}

The algorithm presented in \cite{dawsonProvablySafeTrajectory2020} also provides the gradient of the risk estimate $\nabla_{q_t} \epsilon_\mathcal{O}(q_t)$ with very little overhead. Dawson \textit{et al.} use this gradient to develop a gradient-based chance-constrained trajectory optimization algorithm (considering only uncertainty in the environment). In the following section, we will show how this gradient can also be used in a robust optimization algorithm that considers state uncertainty (i.e. tracking error) in addition to environmental uncertainty, substantially improving the safety of the optimized trajectories in representative scenarios.

\section{Trajectory optimization with uncertain state}\label{state-uncertainty}

In the previous section, we saw how $\epsilon$-shadows can be used to find nominal trajectories with some level of safety in the face of environmental uncertainty. However, due to state uncertainty and tracking error, it is unlikely that these nominal trajectories will remain safe when executed. In this section, we will develop an robust, tractable approximation to the chance-constrained trajectory optimization problem~\eqref{full_ccNLP} that considers both environmental uncertainty and tracking error.

Before presenting our robust optimization approach, it is helpful to define some vocabulary to speak about safety under these two different types of uncertainty.

\begin{definition}[$\delta$-safety]
    A trajectory $q$ is said to be \textbf{$\bm{\delta}$-safe} if the probability of the trajectory colliding with any uncertain obstacle is no greater than $\delta$. That is, the trajectory is $\delta$-safe if $\textrm{Pr}\pn{\bigwedge_{0 \leq t \leq T} \bigwedge_{\mathcal{O} \in E} \textrm{sd}_{\mathcal{O}}(q_t) \geq 0} \leq \delta$.
\end{definition}

In this conception, $\delta$-safety can be a property of either a nominal trajectory $\bar{q}$ or a specific execution of that trajectory $q \sim \cN(\bar{q}, \Sigma_q)$. The $\epsilon$-shadow approach presented in Section~\ref{iros-recap} is concerned with finding $\delta$-safe nominal trajectories when $q = \bar{q}$. It should be clear that $\sum_{0 \leq t \leq T} \sum_{\mathcal{O} \in E} \epsilon_{\mathcal{O}}(\bar{q}_t) \leq \delta$ is a sufficient condition for a nominal trajectory to be $\delta$-safe.

The next definition considers what happens to $\delta$-safe nominal trajectories when they are executed subject to Gaussian tracking error.

\begin{definition}[$\gamma$-robustness]
    If a nominal trajectory is $\delta$-safe, then we say that it is also \textbf{$\bm{\gamma}$-robust} if the probability that an execution $q \sim \cN(\bar{q}, \Sigma_q)$ is also $\delta$-safe is at least $1-\gamma$.
\end{definition}

In other words, a nominal trajectory is $\gamma$-robust if it is $\delta$-safe and also likely (with high probability) to yield executions that are also $\delta$-safe. If a nominal trajectory $\bar{q}$ is $\gamma$-robust, then there is at most probability $\gamma$ that the robot will (at execution time) incur a collision risk greater than $\delta$.

By leveraging the language of $\epsilon$-shadows from Section~\ref{iros-recap}, we see that a sufficient condition for $\gamma$-robustness is
\begin{align}
    \textrm{Pr}\pn{\sum_{0 \leq t \leq T} \sum_{\mathcal{O} \in E} \epsilon_{\mathcal{O}}(q_t) \leq \delta\ \middle|\ q \sim \cN(\bar{q}, \Sigma_q) } \geq 1-\gamma
\end{align}

We can also upper-bound the total risk of collision while executing a $\gamma$-robust, $\delta$-safe trajectory:
\begin{align}
    \textrm{Pr}\pn{\text{collision}} &= 1 - \textrm{Pr}\pn{\neg\ \text{collision}} \\
    &\leq 1 - \textrm{Pr}\pn{\delta \text{-safe}}\textrm{Pr}\pn{\text{collision}\ |\ \delta \text{-safe}} \\
    &\leq 1 - (1-\gamma)(1-\delta) \\
    &\leq \gamma + \delta \label{gamma_delta_upperbound}
\end{align}
Thus, if the user-specified risk tolerance is $\Delta$, then the constraint $\gamma+\delta \leq \Delta$ is sufficient to ensure that a $\gamma$-robust, $\delta$-safe trajectory satisfies the user's risk tolerance.

By combining these sufficient conditions, we can write a conservative approximation of our original chance-constrained optimization problem~\eqref{full_ccNLP}. In this approximation, we incorporate the parameters $\gamma$ and $\delta$ as decision variables, allowing the optimization program to intelligently allocate the overall risk budget between environmental risk (in $\delta$) and tracking error risk (in $\gamma$).

\begin{align}
    \min_{\bar{q}_0, \bar{q}_1, \ldots, \bar{q}_T;\ \delta, \gamma} \quad & f(\bar{q}_0, \bar{q}_1, \ldots, \bar{q}_T) \tag{ccNLP-2}\label{semiapprox_ccNLP}\\
    \text{s.t.} \qquad & \bar{q}_0 \in \mathcal{Q}_{start};\ \bar{q}_T \in \mathcal{Q}_{final} \\
    & \underset{q \sim \cN(\bar{q}, \Sigma_q)}{\textrm{Pr}}\pn{\sum_{0 \leq t \leq T} \sum_{\mathcal{O} \in E} \epsilon_{\mathcal{O}}(q_t) \leq \delta } \geq 1-\gamma \label{semiapprox_prob}\\
    & \gamma + \delta \leq \Delta \\
    & g_i(\bar{q}_0, \bar{q}_1, \ldots, \bar{q}_T) \leq 0;\quad i \in \mathcal{I}
\end{align}

In order to solve problem~\eqref{semiapprox_ccNLP}, we need to convert the constraint on probability in \eqref{semiapprox_prob} with a deterministic constraint. To do this, we draw inspiration both from sequential convex optimization (SCO \cite{schulmanFindingLocallyOptimal2013}), non-convex optimization problems by repeatedly solving a convex approximation, and from convex risk allocation (CRA \cite{blackmoreConvexChanceConstrained2009}), which reduces linear chance constraints to deterministic convex constraints.

In particular, we use the gradient of $\epsilon$-shadow risk estimates to replace the nonlinear chance constraint \eqref{semiapprox_prob} with an approximate linear chance constraint, which we then reduce to a deterministic constraint using techniques from chance-constrained linear programming. This process begins by linearizing the chance constraint \eqref{semiapprox_prob} about the nominal trajectory $\bar{q}$, which reduces it to a linear chance constraint on a single Gaussian variable. Let $e_t$ denote the tracking error $q_t - \bar{q}_t$, drawn from a joint Gaussian distribution $e \sim \cN(0, \Sigma_q)$
\begin{align}
    \textrm{Pr}&\pn{\sum_{0 \leq t \leq T} \sum_{\mathcal{O} \in E} \epsilon_{\mathcal{O}}(q_t) \leq \delta} \\
    &\approx \textrm{Pr}\pn{\sum_{0 \leq t \leq T} \sum_{\mathcal{O} \in E} \epsilon_{\mathcal{O}}(\bar{q}_t) + \nabla\epsilon_{\mathcal{O}}(\bar{q}_t)e_t \leq \delta} \\
    &= \textrm{Pr}\pn{\sum_{0 \leq t \leq T} \sum_{\mathcal{O} \in E} \nabla\epsilon_{\mathcal{O}}(\bar{q}_t)e_t \leq \delta - \sum_{0 \leq t \leq T} \sum_{\mathcal{O} \in E} \epsilon_{\mathcal{O}}(\bar{q}_t)} \\
    &= \textrm{Pr}\pn{z \leq \delta - \sum_{0 \leq t \leq T} \sum_{\mathcal{O} \in E} \epsilon_{\mathcal{O}}(\bar{q}_t) }
\end{align}
where $z$ is a scalar Gaussian random variable given by $z = \sum_{0 \leq t \leq T} \sum_{\mathcal{O} \in E} \nabla\epsilon_{\mathcal{O}}(\bar{q}_t)e_t \sim \cN(0, R^T\Sigma_q R)$. The variance of $z$ is determined by the gradient of the total risk $\sum_{0 \leq t \leq T} \sum_{\mathcal{O} \in E} \epsilon_{\mathcal{O}}(q_t)$ with respect to the trajectory $q$, which we denote as $R^T = \mat{\sum_{\mathcal{O} \in E} \nabla\epsilon_{\mathcal{O}}(\bar{q}_0), \sum_{\mathcal{O} \in E} \nabla\epsilon_{\mathcal{O}}(\bar{q}_1), \ldots, \sum_{\mathcal{O} \in E} \nabla\epsilon_{\mathcal{O}}(\bar{q}_T)}$ (for notational convenience, we concatenate the gradient terms for each timestep $q_t$).

After making this approximation, we are left with a linear chance constraint on a single Gaussian random variable. Conveniently, this probability is given exactly by the CDF of the Gaussian:
\begin{align}
    \textrm{Pr}&\pn{z \leq \delta - \sum_{0 \leq t \leq T} \sum_{\mathcal{O} \in E} \epsilon_{\mathcal{O}}(\bar{q}_t)} \nonumber\\
    &= CDF_{\cN(0, R^T\Sigma_q R)}\pn{\delta - \sum_{0 \leq t \leq T} \sum_{\mathcal{O}_i \in E} \epsilon_{\mathcal{O}_i}(\bar{q}_t)}
\end{align}

This simplification allows us to write a tractable approximation to \eqref{semiapprox_ccNLP}:
\begin{align}
    \min_{\bar{q}_0, \bar{q}_1, \ldots, \bar{q}_T;\ \delta, \gamma} \quad & f(\bar{q}_0, \bar{q}_1, \ldots, \bar{q}_T) \tag{ccNLP-3}\label{approx_ccNLP}\\
    \text{s.t.} \qquad & \bar{q}_0 \in \mathcal{Q}_{start};\ \bar{q}_T \in \mathcal{Q}_{final} \\
    & \underset{\cN(0, R^T\Sigma_q R)}{CDF}\pn{\delta - \sum_{0 \leq t \leq T} \sum_{\mathcal{O}_i \in E} \epsilon_{\mathcal{O}_i}(\bar{q}_t)} \geq 1-\gamma \label{approx_prob}\\
    & \gamma + \delta \leq \Delta \\
    & g_i(\bar{q}_0, \bar{q}_1, \ldots, \bar{q}_T) \leq 0;\quad i \in \mathcal{I}
\end{align}

Since the simplified chance constraint in \eqref{approx_prob} is a differentiable function of the decision variables, we can apply a sequential convex optimization routine (using SNOPT as the underlying SCO solver \cite{gillSNOPTSQPAlgorithm2005}). At each iteration of the optimization, the non-convex constraints in \eqref{approx_prob} are replaced by the approximate convex constraints derived by linearizing \eqref{approx_prob} about the current solution. Because this optimization approach includes the risk allocations $\delta$ and $\gamma$ (representing the amount of the risk budget allocated to environmental risk and tracking error risk, respectively), we call this approach Sequential Convex Optimization with Risk Allocation (SCORA).

\section{Results}\label{results}

To demonstrate the effectiveness of our approach, we will compare the performance of our proposed SCORA trajectory optimization algorithm against three state-of-the-art planners. These planners are:
\begin{enumerate}
    \item ($\epsilon$-opt) Dawson \textit{et al.}'s $\epsilon$-shadow trajectory optimization, which only considers environmental uncertainty \cite{dawsonProvablySafeTrajectory2020},
    \item (p-Chekov) Dai \textit{et al.}'s p-Chekov trajectory optimization algorithm, which only considers tracking error \cite{daiChanceConstrainedMotion2018}, and
    \item (p-Chekov extended) A modified version of the p-Chekov that we extend to handle environmental uncertainty as well as tracking error.\footnote{The original p-Chekov reasons about uncertainty by sampling from the tracking error distribution; we extend this approach by sampling from the distribution of obstacle positions as well. }
\end{enumerate}

Comparison with (1) and (2) will help justify the need for simultaneously considering both tracking error and environmental uncertainty, while comparison with (3) allows us to benchmark against a state-of-the-art planner that considers both of these types of uncertainty. We will compare these three planners in two representative robotics tasks: a parallel parking maneuver with nonlinear vehicle dynamics and a planning problem with a 10-DOF mobile manipulator.

All experiments were conducted on a single core of an Intel  i7-8565U CPU with a clock speed of 1.8 GHz.

\subsection{Parallel parking with nonlinear dynamics}

\begin{figure}[tbp]
    \centering
    \includegraphics[width=0.9\linewidth]{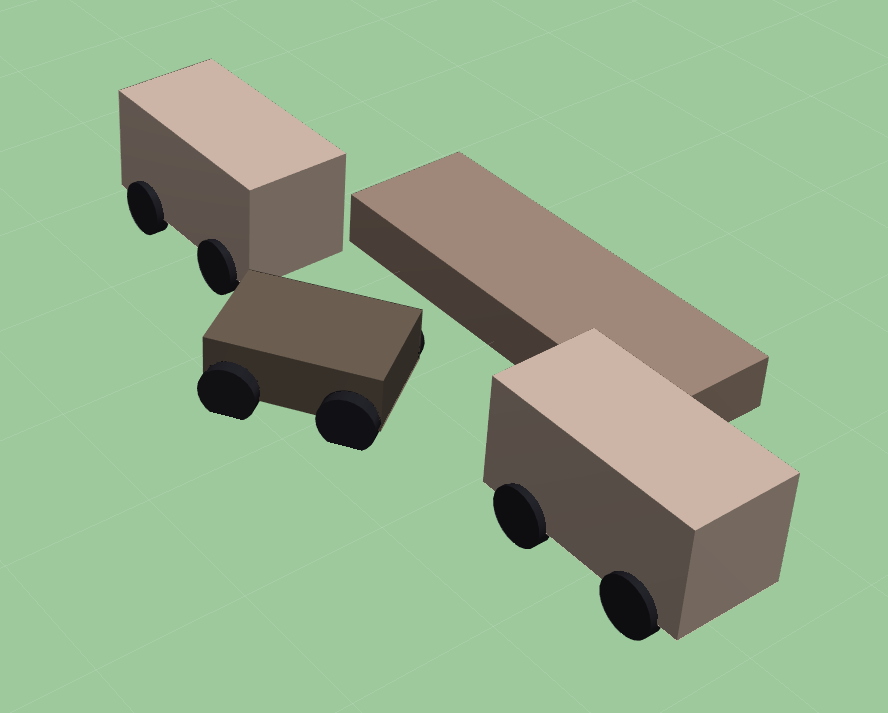}
    \caption{The parallel parking task. The ego vehicle (dark brown) must park between two stationary cars (light brown) without running onto the curb (dark brown).}
    \label{fig:parking_setup}
\end{figure}

In this example, we consider an autonomous vehicle attempting to parallel park between two stationary vehicles, as shown in Figure~\ref{fig:parking_setup}. We model the vehicle dynamics using a discrete time kinematic bicycle model with state $q = [x, y, \theta, v]$ (2D pose and forward velocity, respectively) and control inputs $u=[a, \delta\theta]$ (forward acceleration and steering effort, respectively):
\begin{align}
    x_{t+1} &= x_t + v_t \cos(\theta + \beta) d_t \\
    y_{t+1} &= y_t + v_t \sin(\theta + \beta) d_t \\
    v_{t+1} &= v_t + a_t d_t \\
    \theta_{t+1} &= \theta_t + \frac{v_t}{l_r} d_t \sin{\beta_t} \\
    \beta_t &= \tan^{-1}\pn{\frac{l_r \tan{\delta\theta}}{l_f+l_r}}
\end{align}
where $\beta_t$ is the sideslip angle, $l_r$ and $l_f$ are model parameters for the wheelbase length, and $d_t$ is the discrete time-step duration. The positions of the stationary cars are imperfectly perceived and subject to Gaussian noise with standard deviation $\SI{0.1}{m}$. The ego vehicle's state is subject to tracking error with standard deviation $\SI{0.1}{m}$ in $x$ and $y$ and $0.1$ radians in $\theta$ (SCORA supports tracking error that is correlated between states and across timesteps, but pChekov does not support this and so our experiments are limited to the case of independent noise).

In this example, we use a fixed-time-of-arrival direct transcription optimization scheme, with decision variables for $q_t$ and $u_t$ at each timstep (as well as additional decision variables $\gamma$ and $\delta$ for the risk allocation). We use $16$ time steps of $0.625$ seconds each, and the objective is the sum-squared displacement along the trajectory $0.5 \sum_{t=1}^T ||q_{t} - q_{t+1}||^2$. The overall risk constraint is set to be $0.2$ (this is limited by the p-Chekov algorithms failing to converge for smaller risk constraints; our proposed algorithm successfully planned trajectories for smaller risk bounds).

The results of running each planner on this scenario are shown in Table~\ref{tab:parking_results}. To estimate the true risk of collision for each planned trajectory, we use linear interpolation to up-sample the discrete trajectories to include 100 total waypoints (to check for collision between the optimized waypoints), then ran 1000 independent simulations to check for collisions under environmental uncertainty and tracking error. Runtimes were estimated using an average across 100 executions.

\begin{table}[h]
\caption{Comparison of trajectory optimization algorithms on the parallel parking task (with overall chance constraint $\Delta=0.2$).}\label{tab:parking_results}
\begin{center} 
\begin{tabular}{c||c|c}
Algorithm & Runtime (s)$^1$ & True collision risk$^2$ \\
\hline
$\epsilon$-opt & 0.646 & 22.8\% \\
p-Chekov & 0.120 & 93.0\% \\
p-Chekov extended & 8.326 & 59.1\% \\
SCORA (proposed) & 5.198 & 15.5\% \\
\multicolumn{3}{l}{$^1$ Averaged over $100$ trials.} \\
\multicolumn{3}{l}{$^2$ Averaged over $1,000$ trials (with up-sampled trajectories).}
\end{tabular}
\end{center}
\end{table}

From the results in Table~\ref{tab:parking_results}, we see that SCORA is able to achieve a significantly higher degree of safety than either p-Chekov (which considers tracking error alone) or $\epsilon$-opt (which considers environmental error alone). This demonstrates that considering both tracking error and environmental uncertainty yields a safety benefit beyond that provided by considering either factor alone. It is also interesting to note that comparing the performance of $\epsilon$-opt and p-Chekov suggests that environmental uncertainty has a greater impact on safety than tracking error in this example, since ignoring the effects of environmental uncertainty (as p-Chekov does) results in a higher risk of collision than ignoring tracking error effects (as $\epsilon$-opt does).

We also see that our planner significantly outperforms the extended p-Chekov planner in terms of safety and run-time. This is likely due to the fact that p-Chekov employs a Gauss-Hermite quadrature sampling strategy with only three sampling points to estimate the risk of collision, and this sampling strategy is prone to dramatically underestimating the true risk of collision (increasing the number of sample points would reduce this error at the cost of runtime; we follow the reference implementation in using only three points). As a result, both the original and the extended p-Chekov planners report successfully satisfying the chance constraint when a full Monte Carlo analysis shows the true risk of collision to be unacceptably high. The extent of this overestimate decreases as the magnitude of the state uncertainty decreases, but it nevertheless negatively impacts the performance of these state-of-the-art planners relative to our proposed SCORA algorithm.

Videos of the planned trajectories are included in the supplementary material.

\subsection{10-DOF mobile manipulator}

\begin{figure}[tbp]
    \includegraphics[width=1.0\linewidth]{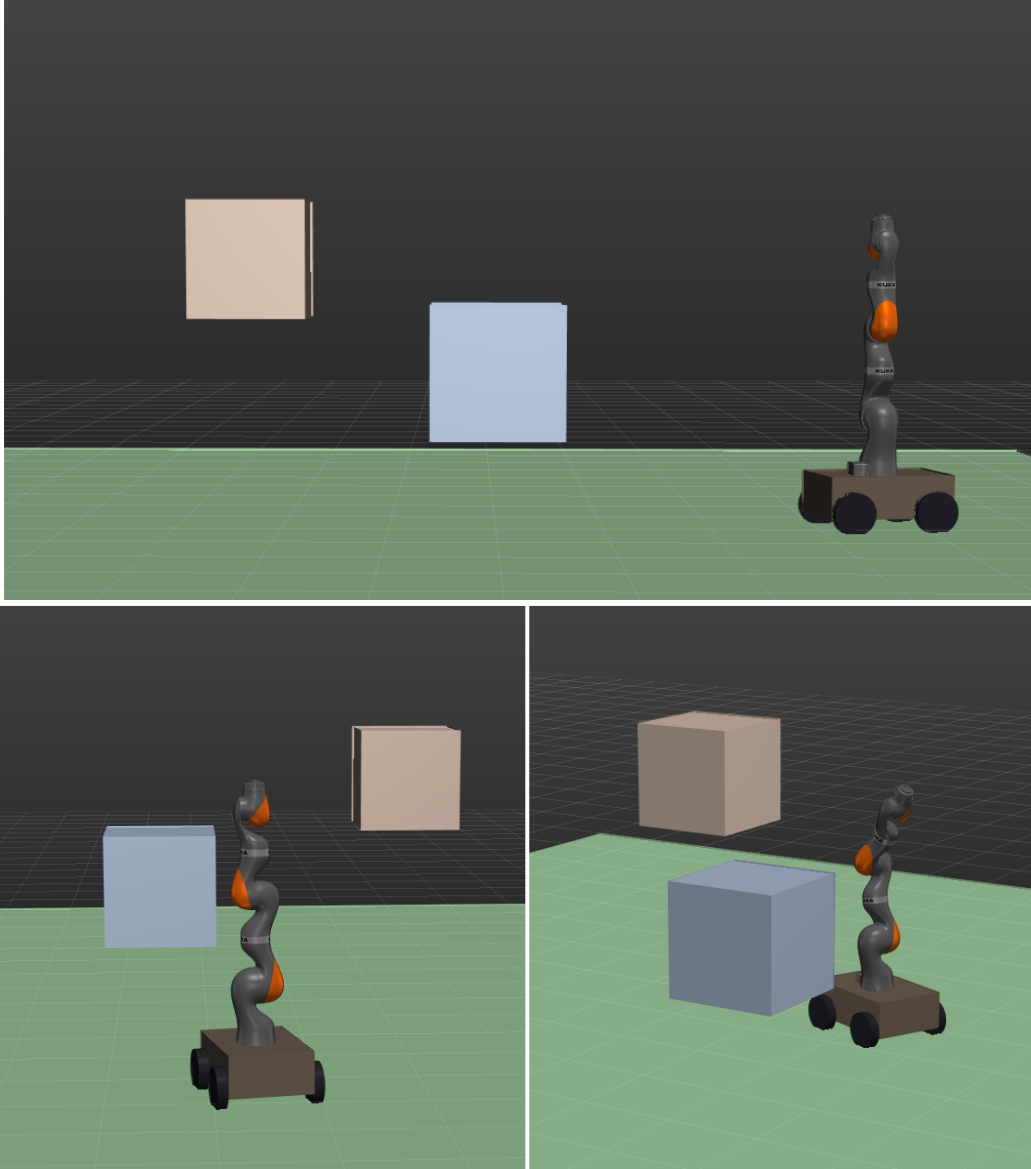}
    \caption{A high-dimensional mobile manipulator navigating around two uncertain obstacles.}
    \label{fig:mobile_arm_setup}
\end{figure}

To demonstrate the scalability of our proposed motion planning algorithm, we consider the mobile-manipulator navigation problem shown in Figure~\ref{fig:mobile_arm_setup}. In this scenario, the mobile arm is tasked with navigating around two uncertain obstacles. Note that although the obstacle geometry is relatively simple, the configuration space of this trajectory planning problem is high-dimensional and complex. The location of each obstacle in this problem is subject to additive Gaussian uncertainty with covariance
\begin{align}
    \Sigma_O = \mat{0.06 & 0.05 & 0.0 \\
                    0.05 & 0.06 & 0.0 \\
                    0.0 & 0.0 & 0.01}
\end{align}
The uncertainty in the state of the robot in this case is dominated by uncertainty in the pose of the mobile base, with tracking error covariance:
\begin{align}
    \Sigma_q = \mat{0.01 I_{2\times 2} & 0 & 0 \\
                    0 & 0.05 & 0 \\
                    0 &  & 0.005 I_{10 \times 10}}
\end{align}
In addition to demonstrating SCORA's scalability for high-degree-of-freedom planning problems, this scenario allows us to demonstrate how the SCORA framework can propagate uncertainty in the pose of the base link to manage collision risk at a distal link.

For these experiments, we use a planning horizon $T=10$ (with timestep $d_t = 0.2$ and a joint collision chance constraint of $\Delta = 0.05$. In addition, we enforce unicycle dynamics on the mobile base:
\begin{align}
    x_{t+1} &= x_t + v_t \cos{\theta_t} d_t \\
    y_{t+1} &= y_t + v_t \sin{\theta_t} d_t
\end{align}
where $v_t$ is the forward velocity at each step, which we constrain $|v_t| \leq 1.0$. The decision variables are the velocity $v_t$ and state $q_t = [x, y, \theta, q_1, q_2, q_3, q_4, q_5, q_6, q_7]_t$ at each timestep (in addition to the risk allocation variables). The objective used in this case is the sum-squared displacement along the trajectory $0.5 \sum_{t=1}^T ||q_{t} - q_{t+1}||^2$.

The results of solving this planning problem with each planner are shown in Table~\ref{tab:mobile_manipulator_results}. In this experiment, neither p-Chekov planner returned a solution in less than $200$ seconds. This is likely because, even with only three sampling points, p-Chekov's Gauss-Hermite quadrature sampling scheme requires $O(3^n)$ collision checks for each outer-loop risk allocation (where $n=10$ is the number of degrees of freedom in the problem), which is prohibitive in this mobile manipulator scenario. For contrast, both $\epsilon$-opt and SCORA avoid this exponential complexity by using $\epsilon$-shadows to upper bound collision risk instead of relying on sampling.

Although the $\epsilon$-opt planner (which only considers external uncertainty) finds a trajectory that is very close to satisfying the chance constraint, we see that SCORA is able to find a solution that is nearly an order of magnitude safer with only a minor ($0.12\%$) increase in the objective. These results show that even in situations where the chance constraint can be satisfied by considering environmental uncertainty alone, it is possible to dramatically increase the safety of planned trajectories --- with relatively little trade-off to the solution cost --- simply by considering tracking error at the same time as environmental uncertainty.

\begin{table}[h]
\caption{Comparison of trajectory optimization algorithms on the mobile manipulator navigation task (with overall chance constraint $\Delta=0.05$).}\label{tab:mobile_manipulator_results}
\begin{center}
\begin{tabular}{c||c|c|c}
Algorithm & Runtime (s)$^1$ & True collision & Optimal cost$^3$ \\
 &  & risk$^2$ & \\
\hline
$\epsilon$-opt & 0.106 & 5.1\% & 1.97 \\
p-Chekov$^4$ & $> 200$ & - & - \\
p-Chekov extended$^4$ & $> 200$ & - & - \\
SCORA (proposed) & 0.198 & 0.7\% & 2.21 \\
\multicolumn{3}{l}{$^1$ Averaged over $100$ trials.} \\
\multicolumn{4}{l}{$^2$ Averaged over $1,000$ trials (with up-sampled trajectories).} \\
\multicolumn{4}{l}{$^3$ Objective is $0.5\sum_{t=1}^T ||q_{t+1} - q_t||^2$.} \\
\multicolumn{4}{l}{$^4$ Both p-Chekov planners failed to return a solution within 200 seconds.}
\end{tabular}
\end{center}
\end{table}

Finally, we note that the absence of strongly nonlinear dynamics in this example greatly simplifies the planning problem, allowing SCORA to run in significantly less time than in the previous example. In fact, in this case SCORA runs only $85\%$ more slowly than $\epsilon$-opt, and both easily clear the approximately $1$ second threshold for real-time path planning.

Videos of the planned trajectories are included in the supplementary material.

\section{Conclusion}

In this paper, we present a risk-aware trajectory optimization algorithm, SCORA, with three core capabilities:
\begin{enumerate}
    \item The ability to manage risk due to uncertainty in obstacles' locations,
    \item The ability to manage risk due to tracking error and other uncertainty in the robot's own state, and
    \item Support for high-dimensional planning problems with non-trivial geometry.
\end{enumerate}
In addition, SCORA enables the user to include arbitrary objective and constraints, allowing it to solve trajectory optimization problems with nonlinear dynamics. Through experiments in simulation, we demonstrate that this algorithm provides significant benefits over state-of-the-art planners in capability, planning time, and safety.

Additionally, our simulation results demonstrate the importance of considering both environmental uncertainty as well as tracking error. In one scenario considering both sources of uncertainty yields significant safety benefits over considering environmental uncertainty alone (with only a minimal tradeoff in trajectory cost). In another scenario, only the SCORA planner (which considers both sources of uncertainty) was able to find trajectories satisfying the chance-constraint.

\section*{Acknowledgments}

This work was sponsored by Airbus SE.

\bibliographystyle{plainnat}
\bibliography{references}

\end{document}